\documentclass[letterpaper, 10 pt, conference]{ieeeconf}  

\IEEEoverridecommandlockouts                              

\usepackage{hyperref}
\usepackage{makecell} 
\usepackage{resizegather}
\usepackage{enumerate} 
\usepackage{times}
\usepackage{epsfig}
\usepackage{graphicx}
\usepackage{amssymb} 
\usepackage{amsmath}
\usepackage[linesnumbered,ruled,vlined]{algorithm2e}
\usepackage{algorithmic}
\usepackage{array}
\usepackage{rotating}
\usepackage{multirow}
\usepackage{csquotes}
\usepackage{stfloats}
\usepackage{color}
\usepackage[dvipsnames]{xcolor}
\usepackage{float}  
\usepackage{etoolbox}
\usepackage{balance}
\usepackage{cite}
\makeatletter
\patchcmd{\@makecaption}
  {\scshape}
  {}
  {}
  {}
\makeatletter
\patchcmd{\@makecaption}
  {\\}
  {.\ }
  {}
  {}
\makeatother
\def\tablename{Table}

\usepackage[font=scriptsize,labelfont=sf,textfont=sf]{subcaption}
\SetKwInput{KwInput}{Input}
\newcolumntype{P}[1]{>{\centering\arraybackslash}p{#1}}

\newcommand{\thickhline}{%
    \noalign {\ifnum 0=`}\fi \hrule height 1pt
    \futurelet \reserved@a \@xhline
}
\title{\LARGE \bf A Framework for Fast Prototyping of\\ Photo-realistic Environments with Multiple Pedestrians}

\author{Sara Casao$^{1}$ \hspace{0.5cm} Andr\'{e}s Otero$^{1}$ \hspace{0.5cm} \'{A}lvaro Serra-Gómez$^{2}$ \hspace{0.5cm} \\ Ana C.~Murillo$^{1}$\hspace{0.5cm} Javier Alonso-Mora$^{2}$\hspace{0.5cm}
 Eduardo Montijano$^{1}$
\thanks{$^{1}$ S. Casao, A. Otero, A.C. Murillo and E. Montijano are with RoPeRt group, at DIIS - I3A, Universidad de Zaragoza, Spain. {\tt\small \{scasao, aotero, emonti, acm\}@unizar.es }}
\thanks{$^{2}$ A. Serra-Gómez, J. Alonso-Mora are with Cognitive Robotics, at TU Delft, The Netherlands. {\tt\small \{A.SerraGomez@ , J.AlonsoMora@\}@tudelft.nl}
}
}

\begin{document}

\maketitle
\thispagestyle{empty}
\pagestyle{empty}

\begin{abstract}
Robotic applications involving people often require advanced perception systems to better understand complex real-world scenarios.
To address this challenge, photo-realistic and physics simulators are gaining popularity as a means of generating accurate data labeling and designing scenarios for evaluating generalization capabilities, e.g., lighting changes, camera movements or different weather conditions.
We develop a photo-realistic framework built on Unreal Engine and AirSim to generate easily  scenarios with pedestrians and mobile robots.
The framework is capable to generate random and customized trajectories for each person and provides up to 50 ready-to-use people models along with an API for their metadata retrieval. 
We demonstrate the usefulness of the proposed framework with a use case of multi-target tracking, a popular problem in real pedestrian scenarios. The notable feature variability in the obtained perception data is presented and evaluated.

\end{abstract}

\section*{Supplementary Material}
The framework code, models and generated datasets are available at \url{https://github.com/saracasao/Pedestrian_Environment}

\section{Introduction}
\label{sec:intro}
Multiple problems addressed in robotics, such as tracking, navigation or mapping, entail the presence of people in their real-world applications~\cite{tallamraju2019activeTrack, cao2019CrowdNav,yue2020mapping, casao2021distributed}.
Performing a systematic evaluation of such methods in environments with humans is a major problem. 
While datasets serve as a traditional option for ranking different approaches under the same testing conditions~\cite{chavdarova2018wildtrack,dendorfer2021mot20, kohl2020mta, kerim2021weather}, their generation entails a resource-intensive process of data gathering and labeling.
Besides, robots move, which requires online testing which makes it challenging to perform a rigorous sensitivity analysis.
Photo-realistic simulators are becoming an increasingly popular tool to overcome these limitations~\cite{devo2021enhancing, chen2020uavsdataset}.
Unfortunately, the generation of complex pedestrian scenarios can still be a time-consuming solution with a steep learning curve, in addition to the tedious work of collecting and blending suitable actors (avatars, movements, textures, etc.).

This work presents a framework to easily generate realistic dynamic scenarios involving mobile robots and numerous pedestrians where researchers can mimic their target application conditions. 
Our framework is built on the photo-realistic and open-source tools Unreal Engine~\cite{unreal} and AirSim~\cite{shah2018airsim}. 
It includes the necessary elements to make the integration smooth and also contains several ready-to-use examples of interest in different robotic applications, namely:

\begin{figure}[t]
\centering
\includegraphics[width=\columnwidth]{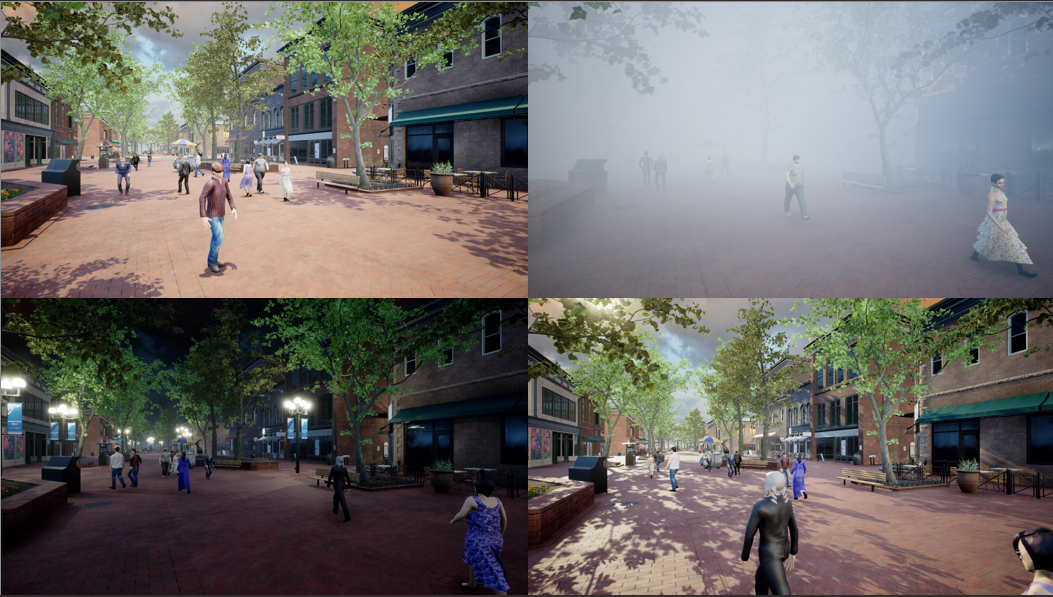}
\caption{Example of the scenarios obtained with the proposed framework. In this example, we use the randomness of pedestrian trajectories to obtain diverse data from the same scenario varying the lighting and weather.}
\label{fig:intro}
\end{figure}

\begin{itemize}
    \item A trajectory plugin implemented in Unreal Engine for a user-friendly definition of paths directly on the environment map. The plugin offers the capability to traverse the created paths in a random or customized fashion. 
    \item A compilation of pedestrian models, created using open-source tools, 
    ready to be used by simply dragging and dropping them on the map. 
    \item A Python API to obtain the environment metadata from AirSim, providing automatic annotations of all the pedestrians present in the scene. 
\end{itemize}
Figure~\ref{fig:intro} shows different examples of the scenes generated with the proposed framework. The randomness of pedestrian trajectories has been leveraged to obtain a large diversity of data within the same environment, exclusively varying light and weather conditions. To demonstrate the usefulness of the developed framework, we adopt the popular use case of multi-target tracking, which consists in determining the position of every person at all times. The evaluation videos are recorded using drones as moving cameras.

\begin{figure*}[t]
\centering
\includegraphics[width=0.99\textwidth]{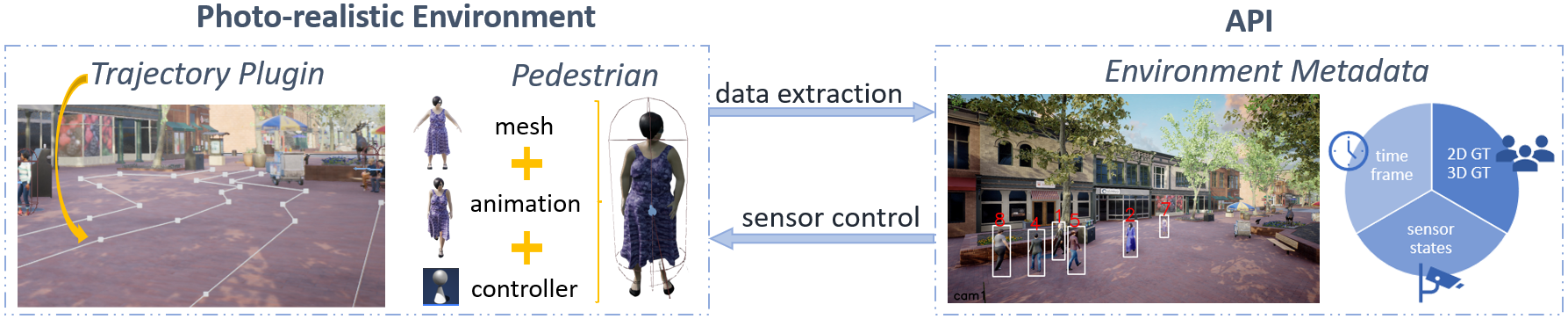}
\caption{Overall structure of the proposed framework. In the photo-realistic environment, the trajectory plugin provides a user-friendly way to define the path to follow for the pedestrian models (white lines). In addition, 50 pedestrian models are ready to be used, where the appearance (mesh), the movement (animation) and the pedestrian behavior (controller) are integrated. Finally, the API extract the environment metadata.}
\label{fig:overall}
\end{figure*}

The rest of the paper is organized as follows. Section~\ref{sec:related} describes the work related to the proposed framework. Section~\ref{sec:framework} gives all the details of the integration of the tools for creating the pedestrian scenario. Section~\ref{sec:experiments} demonstrates the large variability of features that could be easily obtained with the proposed framework and presents the evaluation of multi-target tracking methods with the acquired data. Finally, Section~\ref{conclusions} presents the conclusions of the work.

\section{Related Work}
\label{sec:related}
Most studies that address issues related to autonomous robots concentrate on simulating crowds and individuals by depicting them as points~\cite{grzeskowiak2021crowd}, or they do not offer a wide range of appearance models~\cite{alami2020hateb,favier2021personsimulating}. 
Unfortunately, these works suffer from a lack of perception information, making it highly challenging to create comprehensive solutions that integrate perception and robotic control. 
One of the biggest problems in tackling perception issues is the lack of data to work with, thus, slowing down research progress. Multiple benchmarks have been published focusing on the presence of people in the environment, with one of the most popular being the MOT Challenges. The MOT Challenges goal consists of improving the multi-target tracking task on a single static and moving camera~\cite{leal2015mot15, milan2016mot16, dendorfer2021mot20}. Following this lead, other works, such as~\cite{chandra2019densepeds}, make available new dense pedestrian crowd datasets, e.g., from 2 to 2.7 pedestrians per square meter, in addition to novel approaches for multi-target tracking. Aiming to broaden the posed problem of multi-target tracking to consider other issues like re-identification or consensus, a few works provide benchmarks with multiple overlapping cameras~\cite{berclaz2011EPFL, chavdarova2018wildtrack}. 

In order to alleviate the huge resources cost of gathering and labeling real-world data, the use of photo-realistic simulators has become extremely popular for a wide variety of problems, from navigation~\cite{vorbach2021causal} to cinematography~\cite{pueyo2022cinempc}. 
Some works have leveraged the potential of these simulators to address end-to-end active tracking by training navigation policies based on reinforcement learning and giving the RGB image as the only input~\cite{tallamraju2019activeTrack, luo2019end}.
Regarding the use of simulators for collecting synthetic data to address novel issues, new benchmarks have been published. The detection and tracking of occluded body joints along with a new dataset obtained from the \textit{Grand Theft Auto V (GTA V)} videogame are presented in~\cite{fabbri2018occlusion}. Taking advantage of the same simulator, \cite{kohl2020mta} makes available a benchmark for multi-target tracking in a multi-camera system with and without overlapping cameras, while \cite{kerim2021weather} with \textit{NOVA} focuses on evaluating the robustness of tracking methods under adverse weather conditions. 
Other works, such as~\cite{da2022actionRecognition, zhang2021unrealperson}, tap into the effortlessness of getting labeled data with simulators and propose domain adaptation algorithms from synthetic to real-world data for action recognition and re-identification, respectively. The first one~\cite{da2022actionRecognition}, renders people in \textit{Blender}, whereas the second one~\cite{zhang2021unrealperson} uses \textit{Unreal Engine}, both of them gathering the required animations to create the people models from Mixamo~\cite{mixamo}. 
Previous works on synthetic datasets rarely release the simulator they use, with the exception of~\cite{zhang2021unrealperson} and~\cite{fabbri2018occlusion}. Unfortunately, Mixamo no longer allows the use of its software for any machine learning or artificial intelligent tasks\footnote{https://www.adobe.com/legal/terms.html} and \textit{GTA V} uses only static cameras with no option of working with drones inside the simulation. 
Recent works focus on making the simulator available and flexible to generate datasets~\cite{alvey2021simulated_uav} and test environments~\cite{muller2018sim4cv} for multiple computer vision applications. However, to the best of our knowledge, none of these works provides neither the required implementations to create fast and easily customized pedestrian scenarios nor releases open-source ready-to-use people models.  

Our work leverages the benefits of Unreal Engine and AirSim to generate photo-realistic interactive environments with a focus on pedestrian scenarios,
providing guidelines to generate challenging conditions for current algorithms.
The effort required to develop a photo-realistic environment that enables dealing with any of the problems mentioned above is considerably high. For that reason, we present a framework that significantly reduces the workload for creating scenarios with moving pedestrians. 

\section{Framework}
\label{sec:framework}
The framework presented in this work provides the essential tools to easily develop photo-realistic simulated pedestrian scenarios for robotic-oriented applications. Figure~\ref{fig:overall} shows the overall structure of the developed implementation.
In the photo-realistic environment managed via Unreal Engine, the trajectory plugin tool gives the ability to create paths and control pedestrians to follow them in a random or customized fashion. We provide a collection of pedestrian models ready to use by simply dragging and dropping them into the scene. These models are composed of self-generated meshes integrated with a database of existing
open-source animations, and controlled via the trajectory plugin.
Finally, a Python API developed in AirSim extracts the environment metadata including images, pedestrian information ground truth, sensor states, and timestamps.    
The open-source feature of the framework allows adding new assets, such as other motion models or new meshes to tailor the scene to user preferences. 
In the following subsections, we explain in detail the different modules and the relations between them.

\subsection{Simulation Environment}
Our framework is integrated within Unreal Engine to benefit from the extensive community and a large amount of freely available 3D models.
Moreover, Unreal has already become a key tool within the robotics community, addressing tasks such as autonomous driving~\cite{osinski2020simulation} or exploration with drones~\cite{alvey2021simulated_uav}, making it a suitable choice for the application at hand. 

\subsubsection{World description}
The first element to consider in the simulation is the world to place the pedestrians. For this part, we directly make use of existing environments at the marketplace and the options available within Unreal to configure them, i.e., lighting and weather conditions. For the sake of fast deployment of the proposed framework, the default installation from GitHub already includes one world ready to use. Nevertheless, changing this part is straightforward following the instructions provided by Unreal. 

\subsubsection{Pedestrians}
The photo-realistic simulation of people entails a high complexity due to the high variability of appearance and intricate movements. Consequently, we focus on providing ready-to-use models of pedestrians. However, it is important to highlight that these models can be exchanged with other agents such as animals or vehicles by simply replacing the mesh and animation shown in Fig.~\ref{fig:overall}.
Our framework gathers a set of 50 ready-to-used rigged pedestrian models. These models consist of realistic human characters, encompassing a diverse cast, in terms of gender, ethnicity, height, or clothing, for a truthful simulation of a real-life environment.

The 3D human meshes are produced in \textit{MakeHuman} \cite{makehuman}, an open-source application that allows the mass production of people with random characteristics by taking into account different adjustable rules. To increase the diversity of human models, we use a community gallery included in the application to download \textit{Creative Commons} assets, such as topologies, skin tones or clothes.
The final people models that may not be realistic or appropriate have been manually filtered out.

\begin{figure}[t]
\centering
\includegraphics[width=\columnwidth]{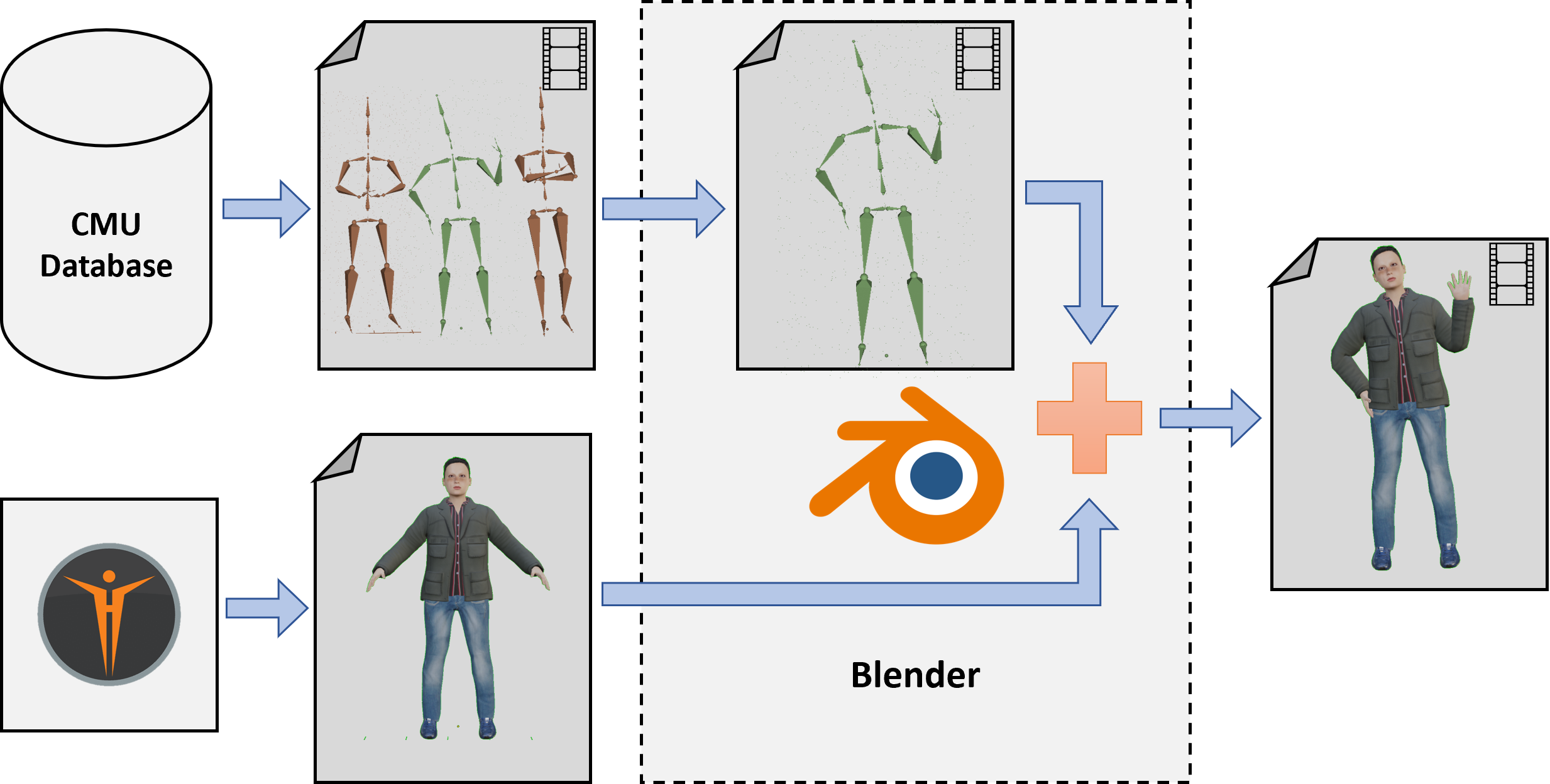}
\caption{Scheme of the characters' models obtention. A subset of animations from the CMU Motion Capture Database is downloaded, and a set of human 3D models are created with MakeHuman. The animations are trimmed to only contain valuable frames and then applied on the skeleton of a model via Blender. Finally, the model performing the animation is exported in Unreal Engine.}
\label{fig:model_process_scheme}
\end{figure}

Next, the pedestrians must mimic the action of walking within the environment, hence, the animations associated with this action have to be blended with the previous meshes.
In our framework, the animations are from the CMU Graphics Lab Motion Capture Dataset \cite{CMUMocapDataset}, which consists of 2605 motion capture segments in different formats.
Focusing on the walking animations, multiple motion capture segments have been assessed to select those most suitable and with sufficient quality for a photo-realistic environment. The final adaptation of these animations to a pedestrian movement has been done in \textit{Blender}, 
where we discard the unrealistic frames in the merge of movement and mesh. Additionally, the frames attributable to the motion capture process are removed, such as the first moments of preparation of the motion capture actor and those at the end of the movement.
Lastly, the models in Blender are exported to Unreal where the integration of all the components, i.e., mesh, animation and controller, is performed. These Unreal pedestrian models are ready to use by simply dragging and dropping them on the world. Figure~\ref{fig:model_process_scheme} shows a scheme of the process.

\subsubsection{Trajectory plugin}
The developed plugin provides multiple trajectory definition options, enabling the effortless creation of scenes with varying levels of complexity.
Firstly, the \textit{continuous path} tool enables the creation of specific trajectories to force characters to walk along them. In more intricate scenarios, i.e., urban areas with countless obstacles, this tool offers a user-friendly fashion to create routes that result in natural, realistic trajectories like walking along the sidewalk. 
The second tool, named \textit{target points}, defines a set of goal positions that pedestrians can reach by crossing any area of the map, being especially useful for open and unconstrained environments.

Both of these tools are exclusively used to define the routes directly in the world. In order to select and walk towards a \textit{continuous path} or a \textit{target point}, the pedestrian models integrate an \textit{AI Controller} that makes the person perform a trajectory in a random or customized manner.   
The former method seeks out goals within an area and selects one randomly, prompting the pedestrian to move in that direction until the goal is reached. Then, the search action is repeated to set the next goal.
Regarding the customized mode, this option is more time-consuming due to requires specifying the goals for each pedestrian. The controller loads the targets tagged with the current pedestrian's name and scrolls the person through them chronologically from oldest to newest. 

Moreover, the developed trajectory plugin includes a user-friendly way of generating scenarios where pedestrians follow the traced path by simply dragging and dropping the \textit{continuous path} or the \textit{target point} directly on the map. This avoids the tedious task of collecting the specific coordinates of the desired trajectory. An example of a different traced route with the \textit{continuous path} can be seen in Figure~\ref{fig:overall}.

\subsection{API for environment metadata capture.}
The API implemented in AirSim performs the image acquisition, the automatic annotation, and the control of the cameras at the scene. AirSim is an open-source simulator built on Unreal Engine that aims to narrow the gap between simulation and reality for the development of autonomous vehicles, particularly drones and cars~\cite{shah2018airsim}.
Different points of view can be used in the framework by defining fixed cameras, using drones as dynamic cameras, or associating cameras with pedestrians to obtain egocentric perspectives.
The data annotation is conducted automatically by isolating the pedestrians from the rest of the environment with a unique key structure embedded in their Unreal identity. Every pedestrian model provided in this work shares this key structure to be easily found. 
Then, pedestrians' ground truth, including instance segmentation, 3D position, and 2D bounding boxes, are gathered in every iteration of metadata acquisition and saved in a JSON file for ease of use. 

\begin{figure}[t]
\centering
\includegraphics[width=0.33\textwidth]{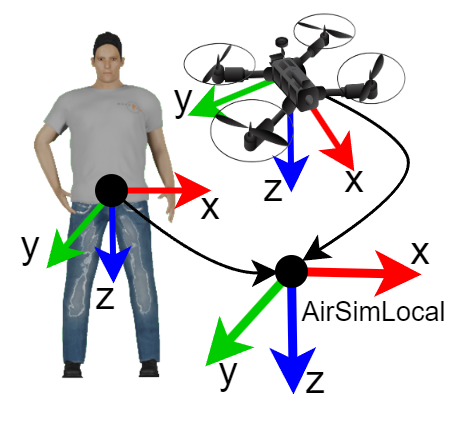}
\caption{Coordinate system of the framework. All acquired metadata is referenced in the \textit{AirSimLocal} coordinate system.}
\label{fig:coordinates_system}
\end{figure}

All the collected information is referenced to the \textit{AirSimLocal} coordinates system, including the dynamic cameras which AirSim references to their initial position, Figure~\ref{fig:coordinates_system}.

\label{sec:experiments}


\begin{figure*}[!b]
\centering
\includegraphics[width=\textwidth]{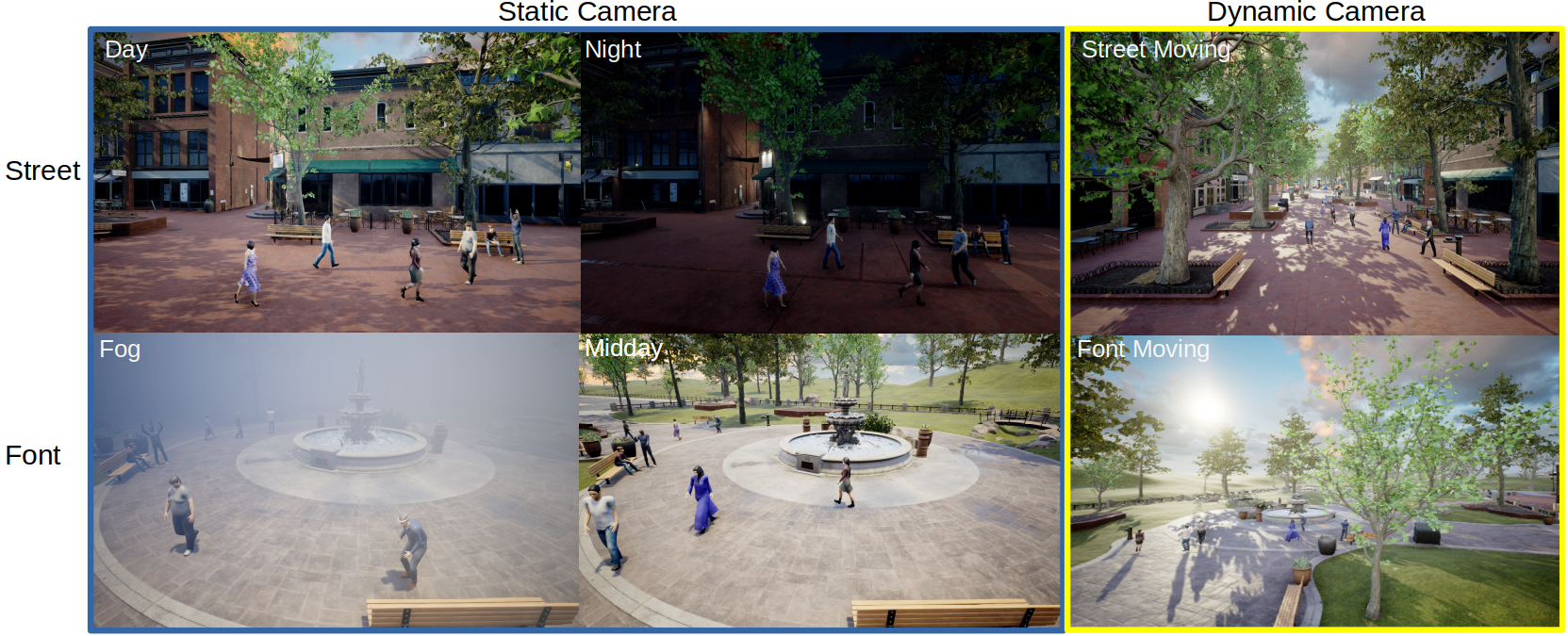}
\caption{Example of the acquired data with the presented framework. Each row is a scene used from the free downloaded map. The first and second columns are  sequences recorded with static cameras, while the third column is acquired from a drone camera.}
\label{fig:img_datasets}
\end{figure*}\section{Experiments}
\label{sec:experiments}

\subsection{Overview}
This section analyzes the multi-target tracking problem as a use case that can benefit from the data generated with the presented framework. 
The high variability of features in the data generated within the simulated environment shows how it significantly aids in a more comprehensive evaluation of generalization compared to using static benchmarks.
The supplementary material demonstrates how the framework is used to generate a new scenario and modify specific features of the scene. 

We select two state-of-the-art tracking methods for the experiments, PHALP~\cite{rajasegaran2022Posetracking} and Tracktor++~\cite{bergmann2019wo}. 
Following standard evaluation procedures for object tracking, evaluation is performed offline on the automatically annotated sequences generated with our framework. To measure the tracker performance we use the CLEAR MOT Metrics~\cite{bernardin2008motmetrics} along with the Identity Metrics~\cite{ristani2016identitymetrics}. The Multiple Object Tracking Accuracy (MOTA) and ID F1 Score (IDF1) quantify two of the main aspects, namely, object coverage and identity.
Note that the goal of the conducted experiments is not to determine the best tracker, but to demonstrate that the diverse features of the acquired set of data allow for isolating and identifying more challenging situations for the tracker. 

\subsection{Acquired Datasets}
This subsection details the sample datasets acquired from our framework to evaluate the selected tracking methods. 

\paragraph*{Map} To obtain datasets simulating real-world scenes, we have downloaded a free photo-realistic map with different areas, such as commercial streets or a park. This map has been released by the Unreal community and we have used it to create two different scenarios: \textit{Font} and \textit{Street}. 

\paragraph*{Recordings} 
Six simulated dynamic scenes of variable length are recorded, getting the labeling automatically from our framework python API. Our reference for the acquired sets of data is the single-view MOT Challenge benchmarks~\cite{leal2015mot15, milan2016mot16, dendorfer2021mot20}, which release short sequences of fixed and moving cameras in real environments. Nevertheless, the photo-realistic simulator has the ability to generate data by combining its resources, such as multiple overlapping and non-overlapping cameras, or using dynamic cameras at the same time as static ones. The supplementary material includes a set of videos where the data acquisition on the proposed framework leverages these combined resources.

Therefore, we include sequences acquired from fixed and moving cameras, which are simulated with drones flying over pedestrians to elude the problem of obstacle avoidance. While the drone moves toward a target position, the API obtains the environment metadata including the extrinsic and intrinsic parameters of the camera. The presented set of data is summarized in Table~\ref{tab:datasets} where each of the datasets is acquired under different light or weather conditions. 
The \textit{Fog}, \textit{Midday} and \textit{Font Moving} datasets use the random trajectory of pedestrians while in \textit{Day}, \textit{Night} and \textit{Day Moving} we create customized trajectories. All datasets except \textit{Font Moving} capture high-quality images ($1920x1080$ resolution). \textit{Font Moving} is captured to represent several challenging features and conditions for tracking algorithms: low-quality images, a moving camera, and a distant camera from the pedestrians.   

\begin{table*}[!t]
\begin{center}
\begin{tabular}{lcccccc} 
\Xhline{3\arrayrulewidth}
\textbf{Dataset} & \textbf{Camera} & \textbf{Length} & \textbf{Resolution} & \textbf{Pedestrian Trajectory} & \textbf{Scene} & \textbf{Description} \\
\hline
\textit{Day} & Static  & 500 & $1920x1080$ & Customized & Street & Pedestrian street at day time\\ 
\textit{Night} & Static & 500 & $1920x1080$ & Customized & Street &  Pedestrian street at night time\\ 
\textit{Fog} & Static & 900 & $1920x1080$ & Random & Font & Fog weather in a small park\\ 
\textit{Street Moving} & Dynamic & 500 & $1920x1080$ & Customized & Street &  Drone as moving camera flying over people \\
\textit{Midday} & Static & 900 & $1920x1080$ & Random & Font & Small park at midday\\ 
\textit{Font Moving} & Dynamic & 600 & $640x480$ & Random & Font & Low quality dataset with a drone as moving camera \\ 
\hline
\Xhline{3\arrayrulewidth}
\end{tabular}
\caption{Details of the sample datasets acquired with the proposed framework for the tracking use case evaluation. }
\label{tab:datasets}
\end{center}
\end{table*}

Figure~\ref{fig:img_datasets} shows images of each dataset where we can appreciate the high variability in terms of light, weather and viewpoint.  

\subsection{Evaluation}
This evaluation focuses on analyzing the benefits that the proposed framework can bring to the tracking algorithms assessment. 
To perform the proposed analysis, we select two methods that address the object tracking task. 

\begin{figure}[!h]
    \centering
    \includegraphics[width=\columnwidth]{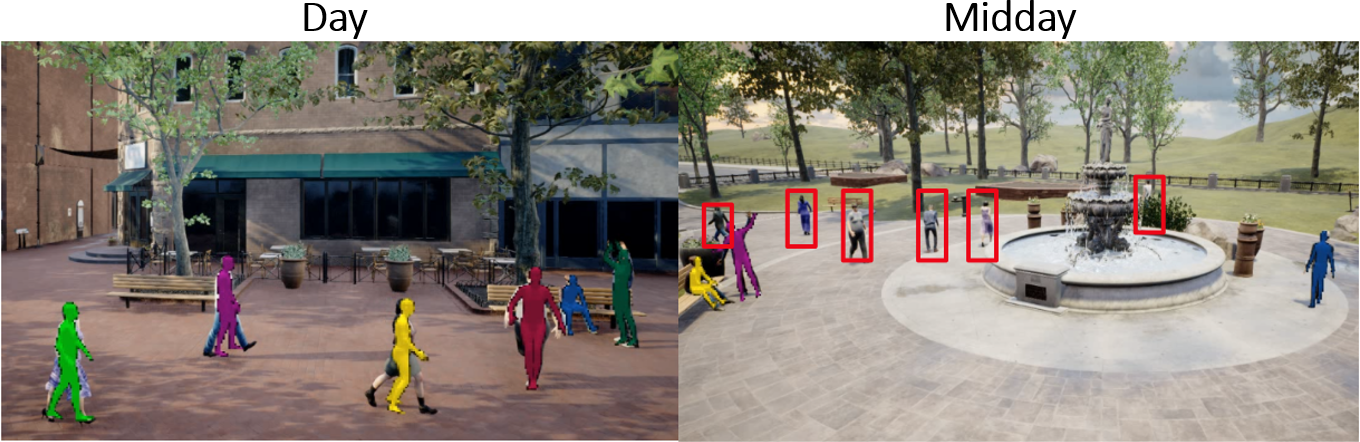}
    \caption{Example of qualitative results with the tracker PHALP. Left: results on \textit{Day} data. Right: results on \textit{Midday} data. The failures (missed people) are highlighted with red bounding boxes. In this case, correspond with people with low resolution.
    }
    \label{fig:phalp}
\end{figure}

The first one is the state-of-the-art PHALP~\cite{rajasegaran2022Posetracking} that tracks people in a single view by predicting the 3D appearance, location and pose. Table~\ref{tab:phalp_results} shows the comparison between their previous results and those obtained with the data acquired with our framework. 
The datasets PoseTrack~\cite{andriluka2018posetrack}, MuPoTS~\cite{mehta2018mupots} and AVA~\cite{gu2018ava} capture a diverse set of high-quality real sequences, including sports, casual interactions and movies. 
The drop in performance in some of our generated sequences, e.g., \textit{Fog}, \textit{Street Moving} and \textit{Midday}, demonstrate that the method is able to deal with space-limited environments where pedestrians are close to the camera but its performance decrease in broad areas up to 11 points in MOTA and IDF1 for static cameras. This effect especially grows when a moving camera is used (\textit{Street Moving}) causing a fall performance of up to 22 points in MOTA. Figure~\ref{fig:phalp} shows a comparison of the qualitative images resulting from this tracker on two scenes. On the left, is an example of the high performance of the method in \textit{Day} dataset, and on the right, is an example on \textit{Midday} where the failures are highlighted with red bounding boxes. 
 
\begin{table}[!h]
\begin{center}
\begin{tabular}{lccc} 
\Xhline{3\arrayrulewidth}
\multirow{2}{*}{Dataset*} &  \multicolumn{3}{c}{PHALP~\cite{rajasegaran2022Posetracking}}\\
& MOTA$\uparrow$ & IDF1$\uparrow$ & IDs $\downarrow$ \\
\hline
PoseTrack~\cite{andriluka2018posetrack} & 58.9 & 76.4 & 541 \\
MuPoTS~\cite{mehta2018mupots} & 66.2 & 81.4 & 22 \\
AVA~\cite{gu2018ava} & - & 62.7 & 227 \\
\hline
Day & 84.4 & 76.2 & 8\\ 
Night & 83 & 82 & 2\\ 
Fog & 48.2 & 52 & 21\\ 
Street Moving & 36.3 & 52.8 & 4\\
Midday & 48.8 & 51.5 & 15\\
\hline
\Xhline{3\arrayrulewidth}
\end{tabular}
\caption{PHALP tracking results in multiple datasets. \textit{PoseTrack}, \textit{MuPoTS} and \textit{AVA} are the benchmarks used for its official evaluation.  \textit{Day}, \textit{Night}, \textit{Fog}, \textit{Street Moving} and \textit{Midday} are the datasets acquired with our framework. We can observe how these sets present a broader range of challenges that enables a more thorough evaluation of the tracker. 
* Sequence \textit{Font Moving}, with poor quality images, is not in this analysis because it makes the algorithm produce unsatisfactory results.}
\label{tab:phalp_results}
\end{center}
\end{table}

\begin{figure}[!t]
    \centering
    \includegraphics[width=\columnwidth]{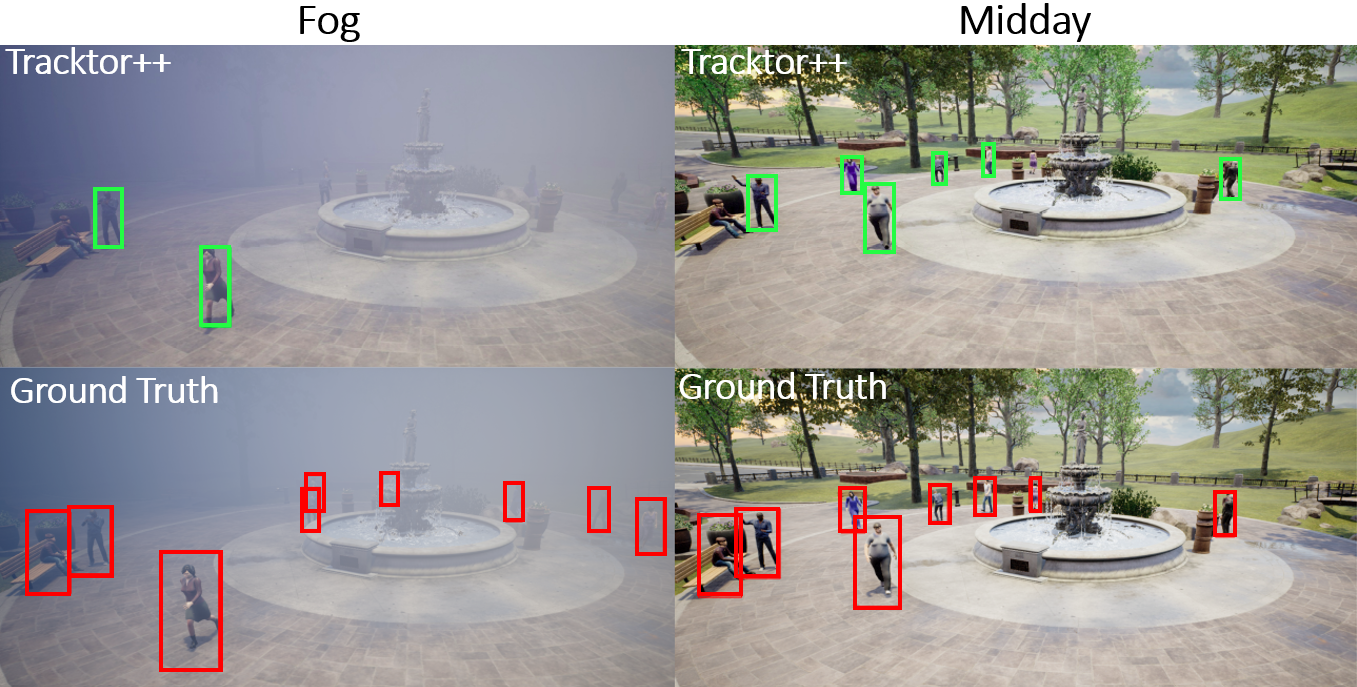}
    \caption{
    Example of qualitative results with the tracker Tracktor++. 
    First row: the bounding boxes depict the tracked pedestrians. Note the difference in performance between \textit{Fog} (left) and \textit{Midday} (right) due to the lack of visibility. Second row shows the ground truth to make the comparison easier for the reader.}
    \label{fig:comparison_vis}
\end{figure}

The second method is the single view tracker Tracktor++~\cite{bergmann2019wo}, whose results on our set of sequences and the MOT Challenges~\cite{leal2015mot15, milan2016mot16} are shown in Table~\ref{tab:tracktor}.
In this case, the lack of visibility is the major cause of failure, Figure\ref{fig:comparison_vis} shows a comparison of the qualitative results obtained with this method between \textit{Fog} (on the left) and \textit{Midday} (on the right). We include a second row with the ground truth to make it easier for the reader to find the pedestrians. In the sunny frame, most people are tracked satisfactorily, however in the foggy image, this precision drops, and only those close to the camera can be tracked. 

We can deduce from the analyzed results that the proposed framework aids in the identification of open problems and the robustness evaluation against the open world.     

\begin{table}[!t]
\begin{center}
\begin{tabular}{lcccc} 
\Xhline{3\arrayrulewidth}
\multirow{2}{*}{Dataset} &  \multicolumn{3}{c}{Tracktor++~\cite{bergmann2019wo}}  \\
&  MOTA $\uparrow$ & IDF1 $\uparrow$ & IDs $\downarrow$\\
\hline
2D MOT2015~\cite{leal2015mot15} & 44.1 & 46.7 &  1318 \\
MOT16~\cite{milan2016mot16} & 54.4 & 52.5 &  682 \\
MOT17~\cite{milan2016mot16} & 53.5 & 52.3 &  2072 \\
\hline
Day & 31.5 & 44.0 & 18 \\ 
Night & 36.9 & 46.4 & 11\\ 
Fog & 23.6 & 32.1 & 20 \\ 
Street Moving & 56.3 & 57.2 & 12\\ 
Midday & 39.3 & 36.9 & 41 \\ 
Font Moving & 3.6 & 9 & 8\\ 
 \hline
 \Xhline{3\arrayrulewidth}
\end{tabular}
\caption{Tracktor++ results in multiple datasets. \textit{2D MOT2015}, \textit{MOT16} and \textit{MOT17} are the benchmarks used for its official evaluation.  \textit{Day}, \textit{Night}, \textit{Fog}, \textit{Street Moving} and \textit{Midday} are the datasets acquired with our framework. 
We can also observe how these sets enable a more thorough evaluation of the tracker.
}
\label{tab:tracktor}
\end{center}
\end{table}

\section{Conclusions}
\label{conclusions}
The framework presented in this work provides all the essential tools to easily develop pedestrian scenarios for robotic-oriented applications. We believe that the released framework will significantly reduce the workload for robotic researchers to develop such environments. The implemented trajectory plugin is a user-friendly tool to create pedestrian paths and control simulated people to follow them in a random or customized fashion. Besides, to alleviate the tedious task of collecting and adapting people models, the framework compiles 50 pedestrian models ready to use by simply dragging and dropping. The provided API automatically gathers and annotates the pedestrians and cameras' metadata. The developed use case highlights how we can easily develop new benchmarks to perform more exhaustive evaluations of applications involving mobile robots and pedestrians, in particular towards robustness to real-world variations.

\section{Acknowledgments}
This work has been supported by  FEDER/Ministerio de Ciencia, Innovación y Universidades – Agencia Estatal de Investigación project PID2021-125514NB-I00,  DGA T45 20R/FSE and the Office of Naval Research Global project ONRG-NICOP-N62909-19-1-2027.

\balance
{
\bibliographystyle{IEEEtran}
\bibliography{egbib}
}
\end{document}